\documentclass[runningheads]{llncs}
\usepackage{graphicx}
\usepackage{amsmath,amssymb} % define this before the line numbering.
\usepackage{color}
\usepackage{bm}
\usepackage{float} 
\usepackage{caption}
\usepackage{subcaption}
\usepackage{multirow}
\usepackage{booktabs}

\usepackage[width=122mm,left=12mm,paperwidth=146mm,height=193mm,top=12mm,paperheight=217mm]{geometry}
\usepackage{cite}
\usepackage{hyperref}
\usepackage{xcolor}
\usepackage{url}

\usepackage{algorithm}
\usepackage{algorithmicx}
\usepackage{algpseudocode}
\usepackage{amsmath} %数学公式
\floatname{algorithm}{Algorithm} %算法
\let\llncssubparagraph\subparagraph
\let\subparagraph\paragraph
\usepackage[compact]{titlesec}
\let\subparagraph\llncssubparagraph
\titlespacing*{\section}{0pt}{2.5ex plus 1ex minus .0ex}{2.0ex plus .0ex}
\titlespacing*{\subsection}{0pt}{1.5ex plus 1ex minus .0ex}{1.5ex plus .0ex}
\titlespacing*{\subsubsection}{0pt}{0.1ex plus 1ex minus .0ex}{0.1ex plus .0ex}
\hypersetup{
  colorlinks=true,
  linkcolor=green!75!black,
  citecolor=green!75!black,
  urlcolor=blue,  % 根据需要调整网址颜色
}

\begin{document}
% \linenumbers
\pagestyle{headings}

\mainmatter
\def\ECCV24SubNumber{***}  % Insert your submission number here

\title{TIER: Text-Image Encoder-based Regression for AIGC Image Quality Assessment} % Replace with your title

\titlerunning{TIER: Text-Image Encoder-based Regression for AIGC Image Quality Assessment}
\authorrunning{J. Yuan, X. Cao, J. Che, Q. Wang, S. Liang, W. Ren, J. Lin, X. Cao}

\author{Jiquan Yuan$^1$, Xinyan Cao$^1$, Jinming Che$^1$, Qinyuan Wang$^1$, \\Sen Liang$^2$, Wei Ren$^1$, 
Jinlong Lin$^1$, \and
Xixin Cao$^1$\thanks{Corresponding author. Email: cxx@ss.pku.edu.cn}}
\institute{$^1$School of Software \& Microelectronics, Peking University, Beijing, China\\
$^2$Microsoft, Beijing, China}

\maketitle

\begin{abstract}
 Recently, AIGC image quality assessment (AIGCIQA), which aims to assess the quality of AI-generated images (AIGIs) from a human perception perspective, has emerged as a new topic in computer vision. Unlike common image quality assessment tasks where images are derived from original ones distorted by noise, blur, and compression, \textit{etc.}, in AIGCIQA tasks, images are typically generated by generative models using text prompts. Considerable efforts have been made in the past years to advance AIGCIQA. However, most existing AIGCIQA methods regress predicted scores directly from individual generated images, overlooking the information contained in the text prompts of these images. This oversight partially limits the performance of these AIGCIQA methods. To address this issue, we propose a text-image encoder-based regression (TIER) framework. Specifically, we process the generated images and their corresponding text prompts as inputs, utilizing a text encoder and an image encoder to extract features from these text prompts and generated images, respectively. To demonstrate the effectiveness of our proposed TIER method, we conduct extensive experiments on several mainstream AIGCIQA databases, including AGIQA-1K, AGIQA-3K, and AIGCIQA2023. The experimental results indicate that our proposed TIER method generally demonstrates superior performance compared to baseline in most cases.
Code will be available at \url{https://github.com/jiquan123/TIER}.
\end{abstract}

\section{Introduction}
Artificial Intelligence Generated Content (AIGC) refers to the process of using AI technology to automatically generate various forms of content including text, images, audio, and video,  \textit{etc.} The core of AIGC involves utilizing machine learning models, particularly deep learning techniques, to understand and mimic the human creative process. The advantages of AIGC include increased efficiency in content creation, reduced costs, and enhanced diversity of creativity, \textit{etc.} With technological advancements, AIGC has been increasingly applied in various fields such as media, advertising, entertainment, and education, \textit{etc.} However, it also poses challenges such as content quality, authenticity, copyright issues, the role and employment impact on human creators, and the risks of being used to create false information or deep fakes, \textit{etc.}

In the realm of image generation, as AI-generative models\cite{r2,bao2023unidiff,dalle,midjourney,nichol2021glide,SDXL22,zhang2023addingcontrolnet,zhou2022lafite} continue to produce images, evaluating the quality of these images has become a significant challenge. Recently, AIGC image quality assessment (AIGCIQA), which aims to assess the quality of AI-generated images (AIGIs) from a human perception perspective, has emerged as a new topic in computer vision. Unlike common image quality assessment tasks where images are derived from original ones distorted by noise, blur, and compression, \textit{etc.}, in AIGCIQA tasks, images are typically generated by generative models using text prompts (Fig.1). Considerable efforts have been made in the past years to advance AIGCIQA. However, most existing AIGCIQA methods\cite{zhang2023perceptual,AGIQA-3K,wang2023aigciqa2023,yuan2023pkui2iqa,yuan2023pscr} regress predicted scores directly from individual generated images, overlooking the information contained in the text prompts of these images. This oversight partially limits the performance of these AIGCIQA methods.

\begin{figure*}[t]
\centering
	\subcaptionbox{}{\includegraphics[height=2cm]{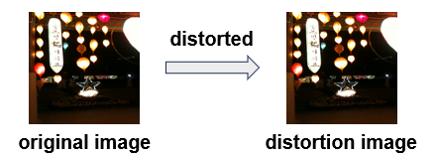}}
	\hfill
	\subcaptionbox{}{\includegraphics[height=2cm]{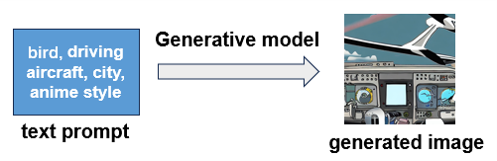}}
\caption{(a) In common image quality assessment tasks, images are derived from original ones distorted by noise, blur, and compression, \textit{etc.} (b) In AIGCIQA tasks, images are typically generated by generative models using text prompts.}
\label{fig:label}
\end{figure*}

To address this issue, we propose a text-image encoder-based regression (TIER) framework. Specifically, we process the generated images and their corresponding text prompts as inputs, utilizing a text encoder and an image encoder to extract features from these text prompts and generated images, respectively. The text encoder employs a text transformer model\cite{Devlin2019BERT,liu2019roberta,lan2019albert,he2020deberta,he2021debertav3} commonly used in natural language processing (NLP), while the image encoder can be a convolutional neural networks (CNN)\cite{he2016resnet,szegedy2015google,szegedy2017inception,simonyan2014vgg} or a vision transformer\cite{dosovitskiy2020VIT}. These extracted text and image features are then concatenated and fed into a regression network to regress predicted scores. Our proposed TIER method builds upon the NR-AIGCIQA method\cite{yuan2023pkui2iqa}, enhancing it by including text prompts as inputs and using a text encoder to extract text features. Consequently, in this study, we utilize the NR-AIGCIQA method as a baseline for comparison with our TIER method. Our primary focus is on examining whether the inclusion of text prompts and the use of a text encoder for feature extraction can yield performance gains, rather than asserting that our method achieves state-of-the-art status.

To demonstrate the effectiveness of our proposed method, we conduct extensive experiments on three mainstream AIGCIQA databases, including AGIQA-1K\cite{zhang2023perceptual}, AGIQA-3K\cite{AGIQA-3K}, and AIGCIQA2023\cite{wang2023aigciqa2023}. The experimental results indicate that our proposed TIER method generally demonstrates superior performance compared to baseline in most cases. Our contributions are summarized as follows:

$\bullet$  We propose a text-image encoder-based regression (TIER) framework. Specifically, we process the generated images and their corresponding text prompts as inputs, utilizing a text encoder and an image encoder to extract features from these text prompts and generated images, respectively.

$\bullet$  We conduct extensive experiments on three mainstream AIGCIQA databases to demonstrate the effectiveness of our proposed TIER framework.

\section{Related Work}
\subsubsection{Image Quality Assessment.}
Over the past few years, many efforts\cite{nr-iqa,r19,r20,cnriqa,iqa1,kim,Li,pan,yan,iqa2, iqa3} have been made to advance image quality assessment (IQA). Recently, AIGC image quality assessment (AIGCIQA), which aims to assess the quality of AI-generated images (AIGIs) from a human perception perspective, has emerged as a new topic in computer vision. As a branch of IQA, researches on AIGCIQA is still relatively scarce. Several dedicated AIGCIQA databases have been established to foster the advancement of AIGCIQA, such as AGIQA-1K\cite{zhang2023perceptual}, AGIQA-3K\cite{AGIQA-3K}, AIGCIQA2023\cite{wang2023aigciqa2023}, and PKU-I2IQA\cite{yuan2023pkui2iqa}, \textit{etc.} Most of these studies utilize the current IQA methods\cite{iqa1, he2016resnet, iqa2, cnriqa, iqa3, simonyan2014vgg} for benchmark experiments. Moreover, Li \textit{et al.}\cite{AGIQA-3K} propose StairReward, significantly enhancing subjective text-to-image alignment evaluation performance. Yuan \textit{et al.}\cite{yuan2023pkui2iqa} introduce FR-AIGCIQA method for full-reference image quality assessment considering both generated images and their corresponding image prompts in predicting quality scores. Yuan \textit{et al.}\cite{yuan2023pscr} propose a patches sampling-based contrastive regression (PSCR) framework to leverage differences among various generated images for learning a better representation space. By selecting exemplar AIGIs as references, they also overcome the limitations of previous models that could not utilize reference images on the no-reference image databases. However, most of them regress predicted scores directly from individual generated images, overlooking the information contained in the text prompts of these images. In this paper, We propose a text-image encoder-based regression (TIER) framework. Specifically, we process the generated images and their corresponding text prompts as inputs, utilizing a text encoder and an image encoder to extract features from these text prompts and generated images, respectively.

\subsubsection{Text Encoder.}
Text encoder plays a pivotal role in the fields of artificial intelligence (AI) and natural language processing (NLP). In the early stages of NLP, the most common text representation methods were the Bag of Words (BoW) model and the Term Frequency-Inverse Document Frequency (TF-IDF) model. These methods, simple yet effective, are primarily used for text classification and search engines. Around 2010, Tomas \textit{et al.}\cite{Mikolov2013word2vec} from Google introduce a word vector model named Word2Vec, capable of capturing semantic relationships between words. Each word is transformed into a fixed-size dense vector, with similar meaning words positioned closely in the vector space. In 2018, context-aware word vector models like Embeddings from Language Models (ELMo)\cite{sarzynska2021ELMo} and Generative Pretrained Transformer (GPT)\cite{Radford2018gpt} emerge. These models generate richer, more dynamic word vectors through bidirectional context or generative pretraining. Also in 2018, Google release Bidirectional Encoder Representations from Transformers (BERT)\cite{Devlin2019BERT}, employing the Transformer architecture and marking the first truly bidirectional text encoder. BERT\cite{Devlin2019BERT} and its derivatives (such as RoBERTa\cite{liu2019roberta}, ALBERT\cite{lan2019albert}, and DeBERTa\cite{he2020deberta}) have significantly improved performance in various NLP tasks. In this paper, we choose BERT\cite{Devlin2019BERT} as the text encoder for extracting features from text prompts.

\subsubsection{Image Encoder.}
Image encoder is a critical component in the field of computer vision, tasked with converting image data into a format that is easier to analyze and process. This conversion typically involves extracting features from raw pixel data for various image understanding tasks, such as image classification, object detection, and image segmentation. Early image encoders relies on various image processing algorithms, like edge detection, Harris Corner Detector, SIFT, and HOG, \textit{etc.} These methods extract features from images using manually designed algorithms, lacking generality and flexibility. After 2012, the advent of deep learning, especially Convolutional Neural Networks (CNNs)\cite{lenet,alexnet,simonyan2014vgg,szegedy2015google,he2016resnet,szegedy2017inception}, mark a significant shift in image encoders. CNNs can automatically learn complex image features from large volumes of labeled data. Representative network architectures include LeNet\cite{lenet}, AlexNet\cite{alexnet}, VGG\cite{simonyan2014vgg}, GoogleNet\cite{szegedy2015google}, and ResNet\cite{he2016resnet}, \textit{etc.}. Around 2014, with the development of transfer learning\cite{transferlearning} and pretrained models, training models on large datasets like ImageNet\cite{russakovsky2015imagenet} and then transferring them to specific tasks allowed for improved performance even with limited data. After 2017, with the rise of attention mechanisms and Transformer architectures\cite{vaswani2017attention}, transformer-based image encoders such as ViT\cite{dosovitskiy2020VIT} emerge. These allow the model to focus more on key parts of an image, particularly excelling in tasks requiring global understanding. In our proposed TIER method, we select several backbone network models that pre-trained on the ImageNet\cite{russakovsky2015imagenet} as image encoders for feature extraction from input images including ResNet18\cite{he2016resnet}, ResNet50\cite{he2016resnet}, InceptionV4\cite{szegedy2017inception}.

\section{Approach}
The overall framework of our method is illustrated in Fig.2. We will describe our method in detail as follows.

\subsection{Problem Formulation}
For a given AI-generated image $I_g$ with a score label $s$ and its corresponding text prompt $T_{I_g}$, we first extract text features from the text prompt $T_{I_g}$ and image features from the generated image $I_g$ utilizing a text encoder and an image encoder, respectively. Subsequently, we fuse these text features and image features by concatenation, and feed them into a regression network to regress predicted scores. This process can be represented as:

\begin{align}
   \hat{s} = R_\theta (\text{concat}(F_{w_1}^T(T_{I_g}), F_{w_2}^I(I_g)))  \label{Eq.4} 
\end{align}

where $R_\theta$ denotes the regression network with parameters $\theta$, $F_{w_1}^T$  represents the text encoder with parameters $w_1$, and $F_{w_2}^I$ represents the image encoder with parameters $w_2$.

\begin{figure}[t]
\centering
\includegraphics[width=12cm]{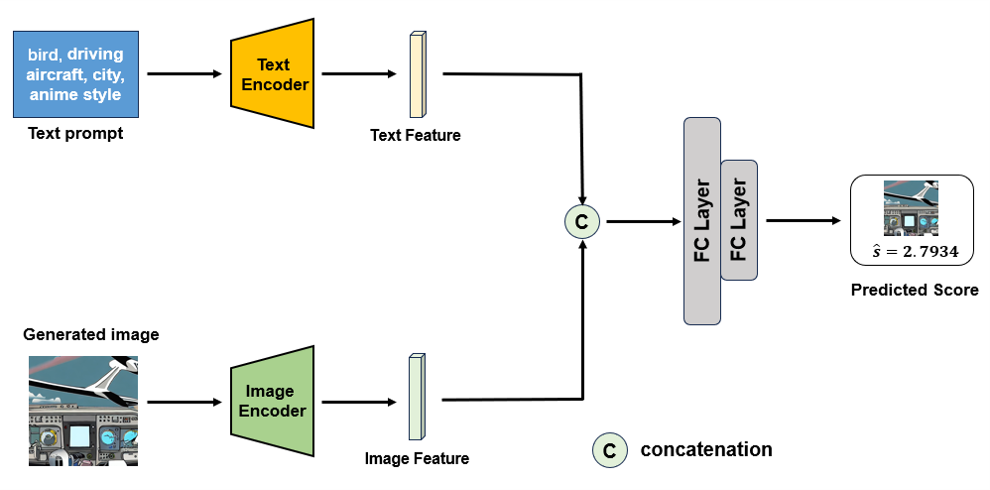} \label{1}
\caption{The pipeline of our proposed text-image encoder-based regression (TIER) framework. We process the generated images and their corresponding text prompts as inputs, utilizing a text encoder and an image encoder to extract features from these text prompts and generated images, respectively. The text encoder employs a text transformer model commonly used in natural language processing (NLP), while the image encoder can be a convolutional neural networks (CNN) or a vision transformer. These extracted text and image features are then concatenated and fed into a regression network to regress predicted scores.}
\end{figure}

\subsection{TIER}
Our proposed TIER method builds upon the NR-AIGCIQA method\cite{yuan2023pkui2iqa}, enhancing it by including text prompts as inputs and using a text encoder to extract text features. It primarily comprises three components: text encoder, image encoder, and regression network. Here is a brief introduction to these three parts.

\subsubsection{Text Encoder.} 
Text encoder plays a pivotal role in the fields of Artificial Intelligence (AI) and natural language processing (NLP). Its primary function is to transform natural language (such as sentences, paragraphs, or entire documents) into numerical representations that machines can understand and process. This process typically involves encoding words, phrases, or sentences into vectors, which capture the semantic and syntactic characteristics of the original text. In this paper, we adopt BERT\cite{Devlin2019BERT} as the text encoder to extract features from text prompts.

\subsubsection{Image Encoder.}
Many classical image quality assessment (IQA) methods initially employ manual feature extraction techniques; however, with the rapid advancement of CNNs\cite{lenet,alexnet,simonyan2014vgg,szegedy2015google,he2016resnet,szegedy2017inception}, feature extraction methods based on deep learning have significantly enhanced performance. Unlike traditional manual feature extraction methods that rely on empiricism to extract features from images, deep learning-based feature extraction methods are data-driven and can capture more abstract and higher-level semantic features from images. In our proposed TIER method, we select several backbone network models pre-trained on the ImageNet\cite{russakovsky2015imagenet} as Image Encoders for feature extraction from input images including ResNet18\cite{he2016resnet}, ResNet50\cite{he2016resnet}, InceptionV4\cite{szegedy2017inception}.

\subsubsection{Regression Network.}
For the fused text-image features with a feature dimension of (B, D), we employ a score regression network composed of two fully connected layers with dimensions $D \times \frac{D}{2}$  and  $\frac{D}{2} \times 1$ to regress the predictd score $\hat{s}$.

\subsubsection{Loss Function.}
We optimize the parameters of the text encoder, image encoder, and the score regression network by minimizing the mean squared error between the predicted score $\hat{s}$ and the ground-truth score $s$:
\begin{align}
   L(\theta, w_1, w_2 | I_g, T_{I_g}) = {MSE}(\hat{s}, s) \label{Eq.4} 
\end{align}

The parameters $\theta$, $w_1$, and $w_2$ correspond to the parameters of the regression network, text encoder, and image encoder, respectively.

\section{Experiment}

\subsection{Datasets}
We perform experiments on three mainstream AIGCIQA benchmarks including AGIQA-1K\cite{zhang2023perceptual}, AGIQA-3K\cite{AGIQA-3K} and AIGCIQA2023\cite{wang2023aigciqa2023}.

\subsubsection{AGIQA-1K.}
The AGIQA-1K database contains 1080 AIGIs generated by two Text-to-Image models\cite{r2} stable-inpainting-v1 and stable-diffusion-v2. To ensure content diversity and catch up with the popular trends, the authors use the hot keywords from the PNGIMG website for AIGIs generation. The generated images are of bird, cat, Batman, kid, man, and woman, \textit{etc.}

\subsubsection{AGIQA-3K.}
The AGIQA-3K database contains 2982 AIGIs generated by six Text-to-Image models including GLIDE\cite{nichol2021glide}, Stable Diffusion V1.5\cite{r2}, Stable Diffusion XL2.2\cite{SDXL22}, Midjourney\cite{midjourney}, AttnGAN\cite{xu2018attngan} and DALLE2\cite{dalle}. This is the first database that covers AIGIs from GAN/auto regression/diffusion-based model altogether.

\subsubsection{AIGCIQA2023.}
The AIGCIQA2023 database contains 2400 AIGIs generated by six of the latest Text-to-Image models including Glide\cite{nichol2021glide}, Lafite\cite{zhou2022lafite}, DALLE\cite{dalle}, Stable-diffusion\cite{r2}, Unidiffusion\cite{bao2023unidiff}, Controlnet\cite{zhang2023addingcontrolnet}. It is constructed by collecting 100 text prompts from PartiPrompts, encompassing 10 scene categories and 10 challenge categories. For each prompt, four distinct images are randomly generated from six generative models, resulting in a comprehensive dataset of 2400 AIGIs (4 images $\times$ 6 models $\times$ 100 prompts), each corresponding to one of the 100 prompts. 

\begin{figure*}[t]
\centering
	\subcaptionbox{}{\includegraphics[width = 4cm]{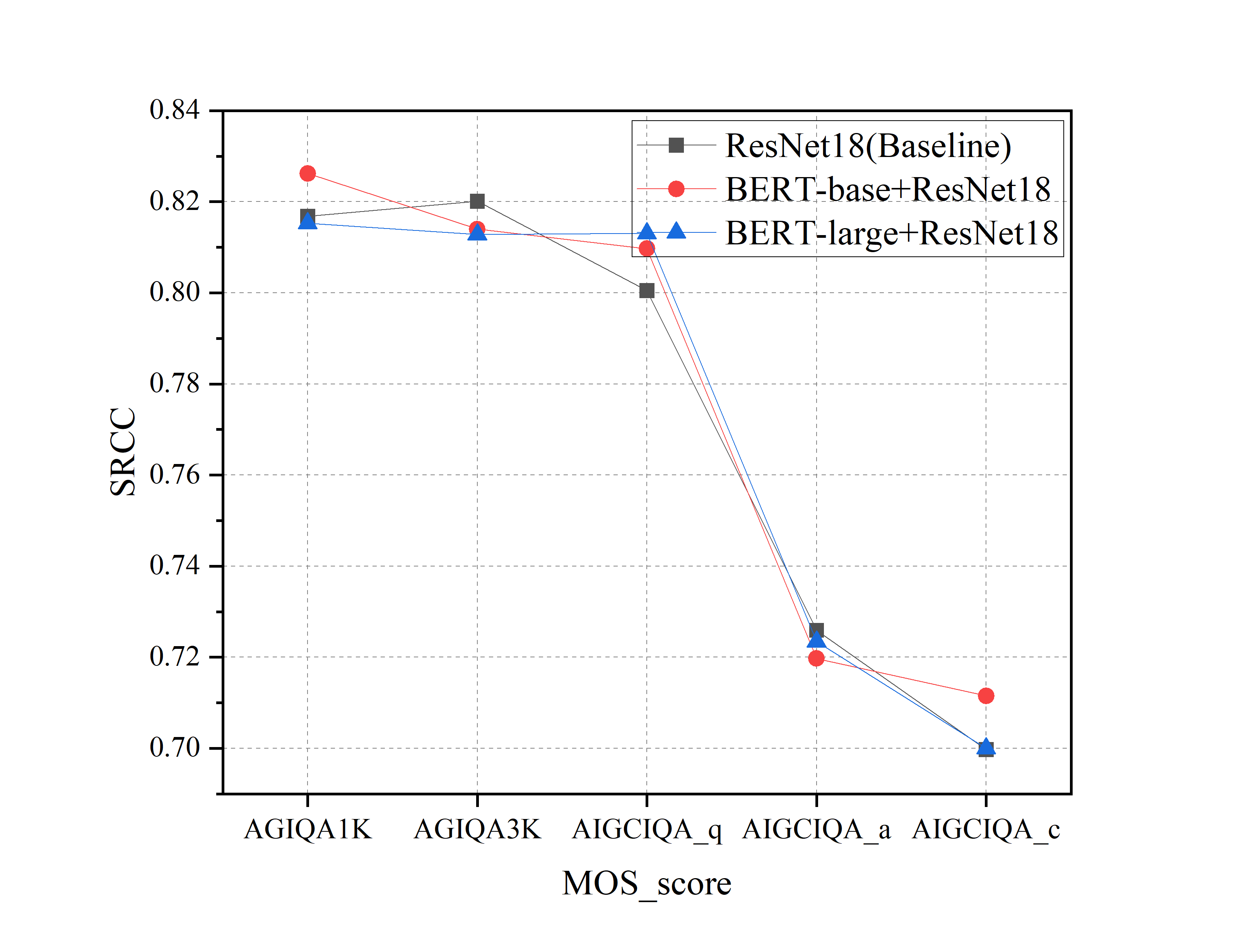}}
	\hfill
	\subcaptionbox{}{\includegraphics[width = 4cm]{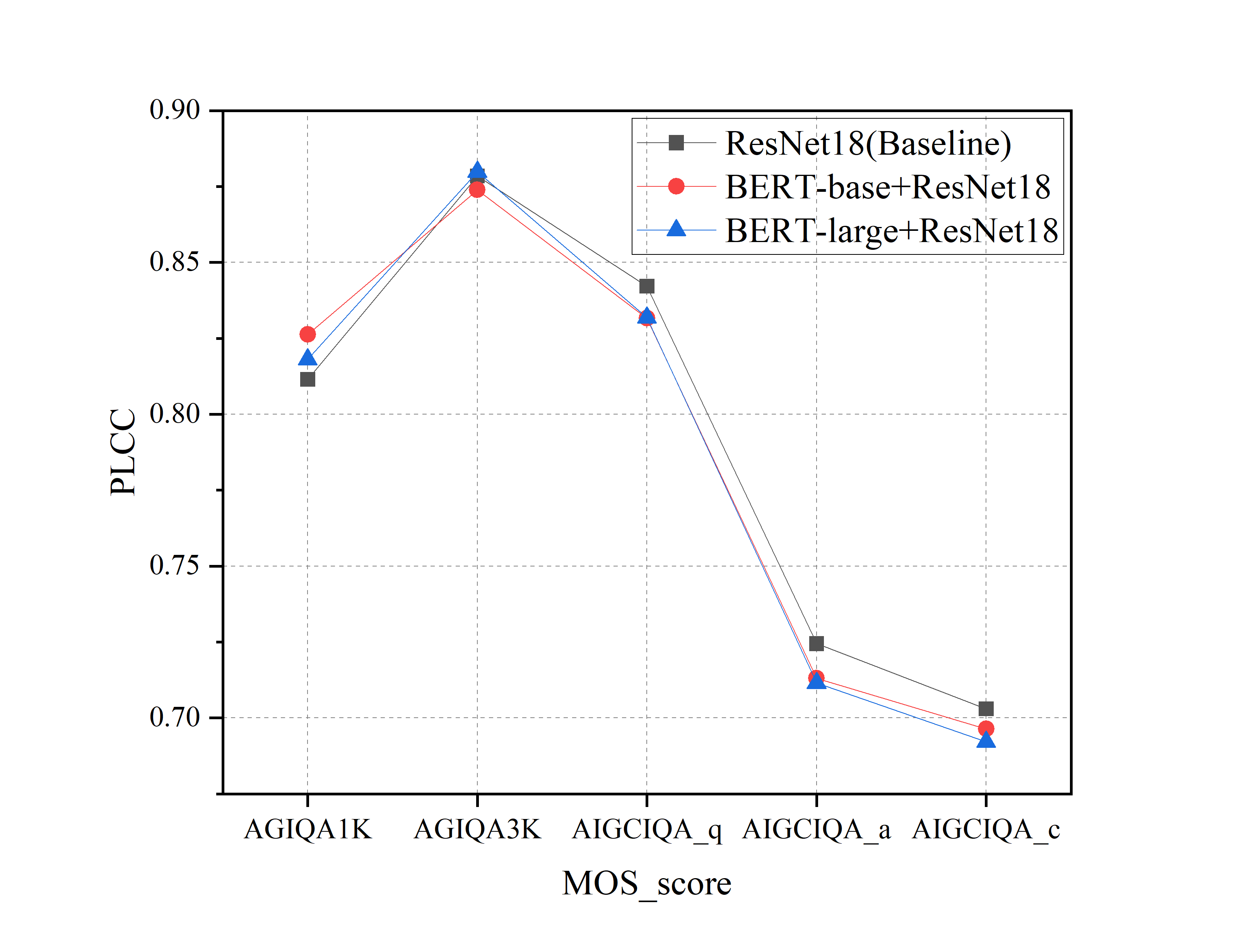}}
	\hfill
	\subcaptionbox{}{\includegraphics[width = 4cm]{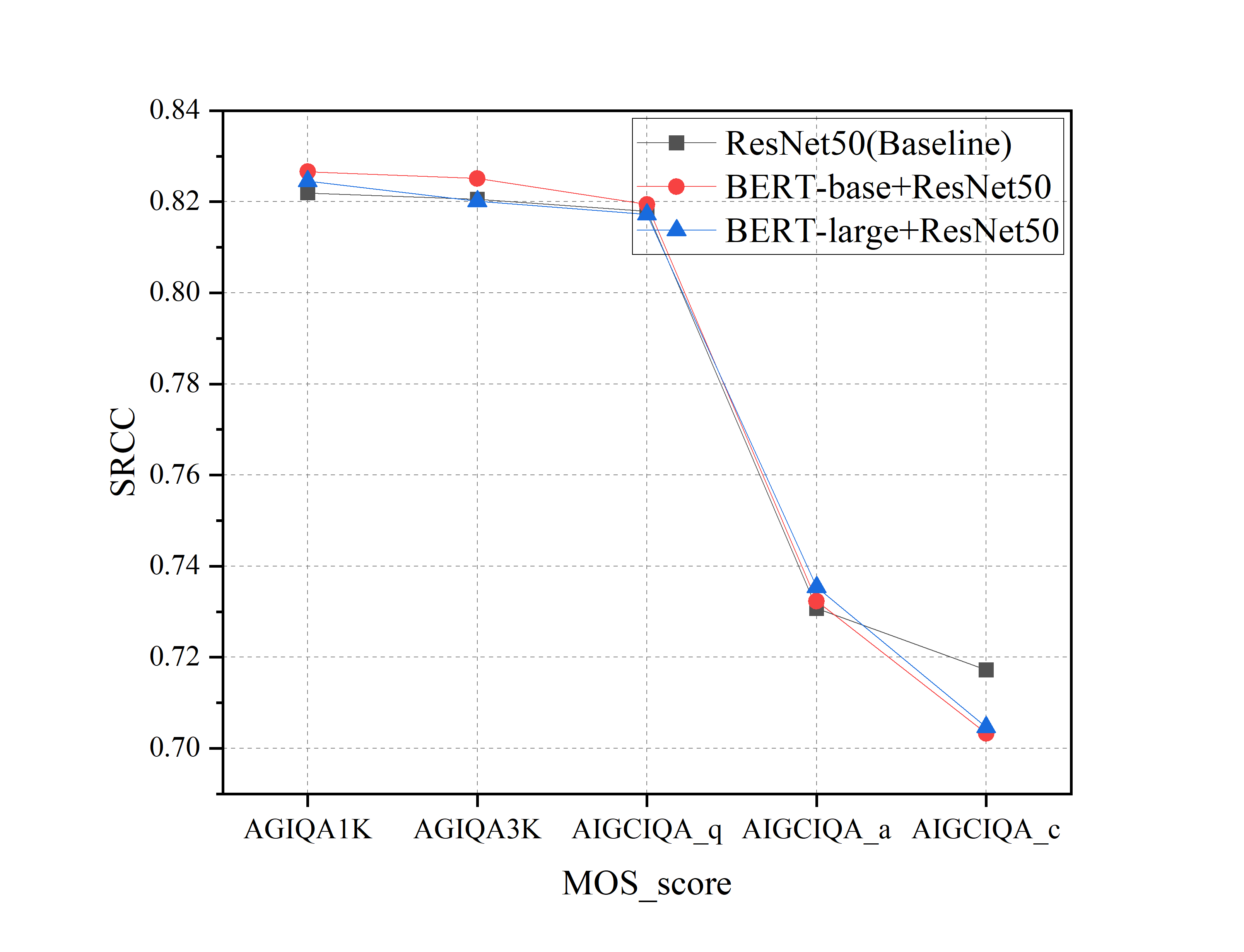}} 
        \hfill
	\subcaptionbox{}{\includegraphics[width = 4cm]{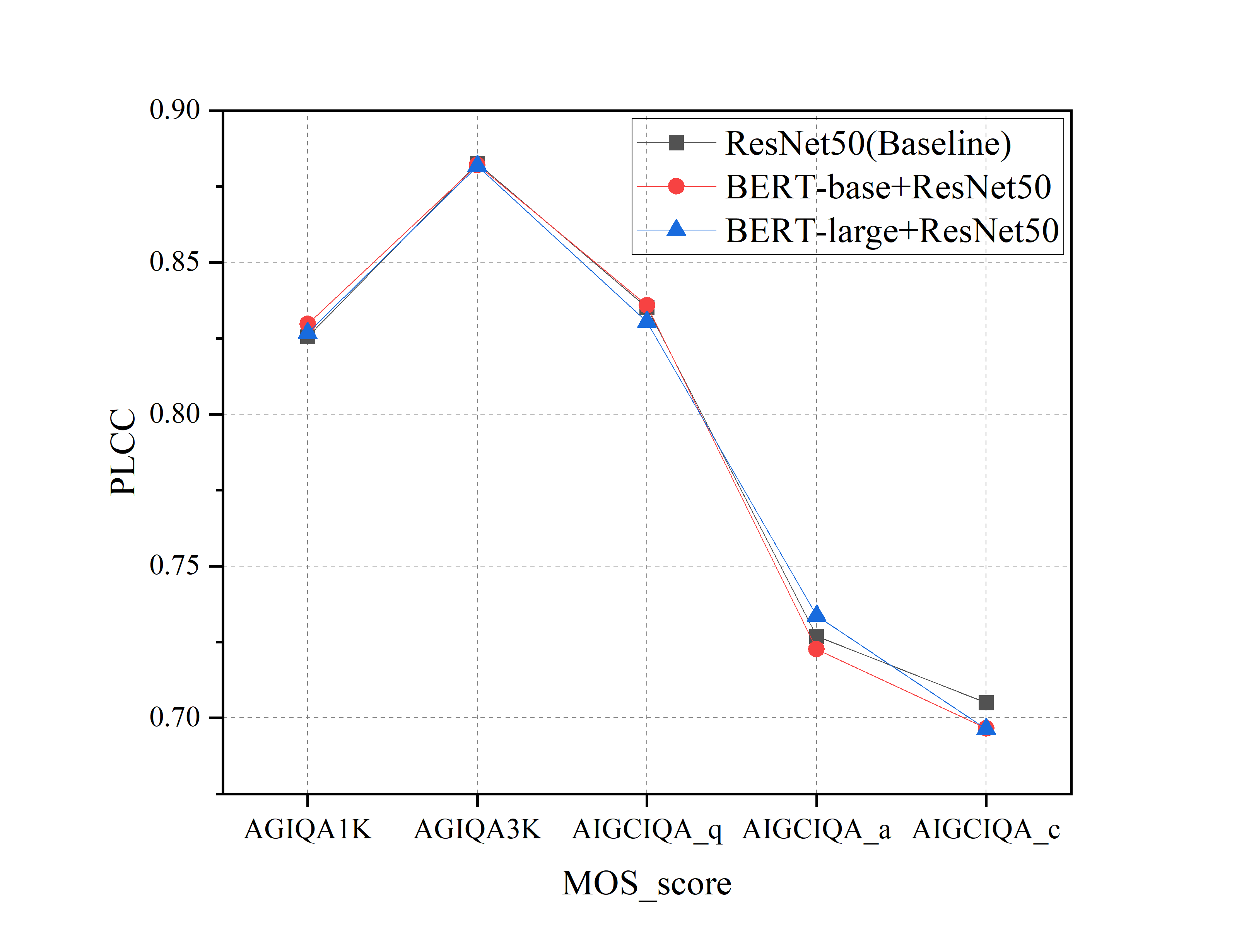}}
	\hfill
	\subcaptionbox{}{\includegraphics[width = 4cm]{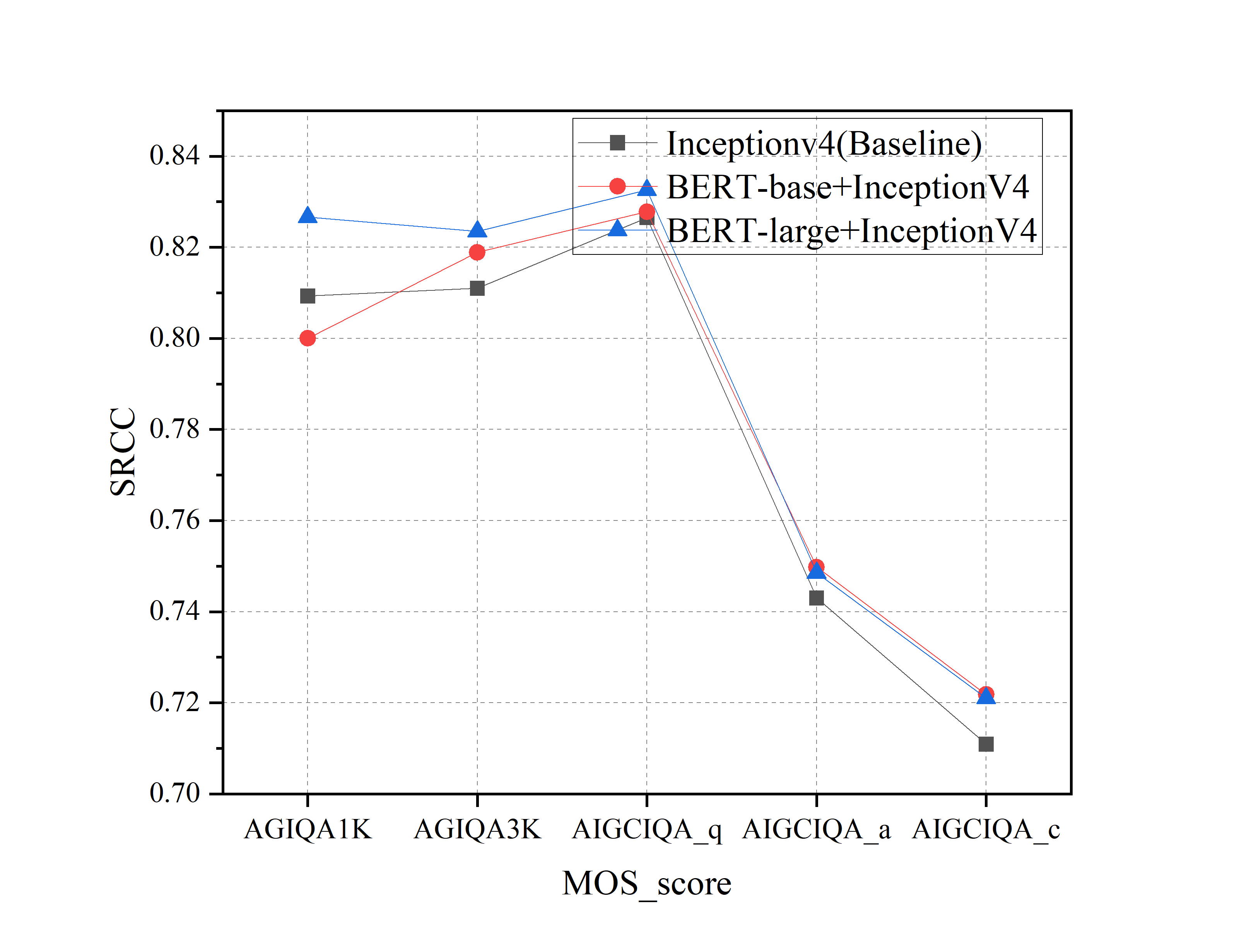}} 
        \hfill
	\subcaptionbox{}{\includegraphics[width = 4cm]{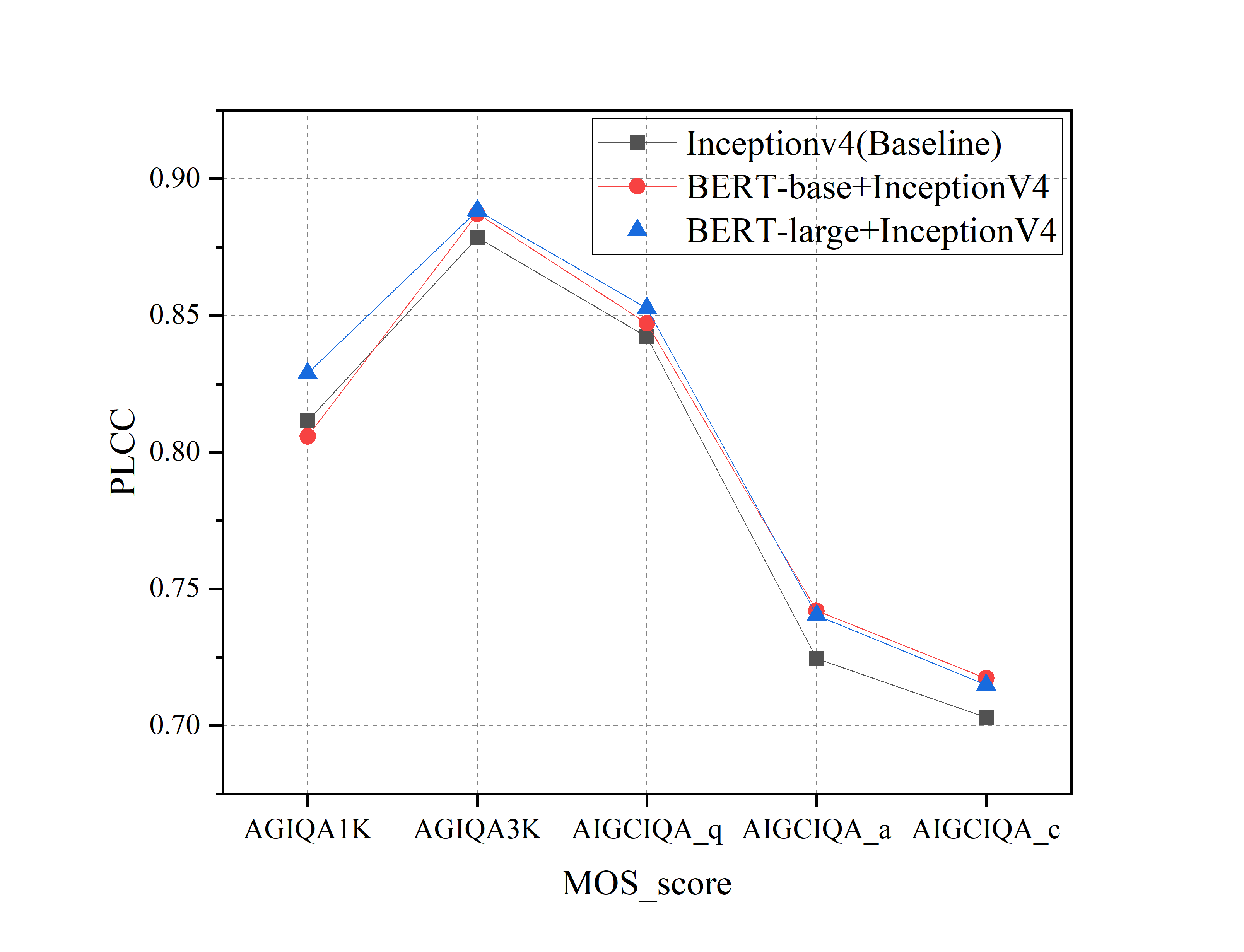}} 
\caption{Comparisons of our proposed TIER method utilizing different Text encoders and Image encoders with baseline on three mainstream AIGCIQA databases.}
\label{fig:label}
\end{figure*}

\subsection{Evaluation Criteria}
In this paper, We employ the Spearman Rank Correlation Coefficient (SRCC) and Pearson Linear Correlation Coefficient (PLCC) as evaluation metrics to evaluate the performance of our methods.

The SRCC is defined as follows:
\begin{align}
   \text{SRCC} = 1 - \frac{6 \sum_{i=1}^{N} d_i^2}{N(N^2 - 1)} \label{Eq.4} 
\end{align}

Here, $N$ represents the number of test images, and $d_i$ denotes the difference in ranking between the true quality scores and the predicted quality scores for the $i_{th}$ test image.

The PLCC is defined as follows:
\begin{align}
    \text{PLCC} = \frac{\sum_{i=1}^{N}(si - \mu_{s_i})(\hat{s}_i - \hat{\mu}_{s_i})}{\sqrt{\sum_{i=1}^{N}(s_i - \mu_{s_i})^2 \sum_{i=1}^{N}(\hat{s}_i - \hat{\mu}_{s_i})^2}}\label{Eq.4} 
\end{align}

Here, $s_i$ and $\hat{s}_i$ represent the true and predicted quality scores, respectively, for the $i_{th}$ image. $\mu_{s_i}$ and $\hat{\mu}_{s_i}$ are their respective means, and $N$ is the number of test images.
Both SRCC and PLCC are range between $-1$ and $1$. A positive
value means a positive correlation. Otherwise, the opposite. A larger value indicates a better performance.

\begin{table}[t]
\centering
\caption{Comparisons of performance with baseline on three mainstream AIGCIQA databases. \textcolor{blue}{$\uparrow$}  indicate that our proposed TIER method demonstrates superior performance compared to baseline.}
\label{tab:overall_table}

\begin{subtable}{1.\linewidth}
\centering
\caption{Comparisons of performance with baseline on the AGIQA-1K database and AGIQA-3K database.}
\begin{tabular}{@{}l|cc|cc@{}}
\toprule
\multirow{3}{*}{Start-index} & \multicolumn{2}{c|}{AGIQA-1K} & \multicolumn{2}{c}{AGIQA-3K}\\
\cline{2-5}
& \multicolumn{2}{c|}{MOS} & \multicolumn{2}{c}{MOS\_quality}  \\
\cline{2-5}
& SRCC & PLCC &  SRCC & PLCC\\
\hline
ResNet18(Baseline)    & 0.8168       & 0.8249       & 0.8201       & 0.8795       \\
ResNet50(Baseline)     & 0.8219       & 0.8255       & 0.8205       & 0.8826       \\
InceptionV4(Baseline)  & 0.8093       & 0.8115       & 0.8110       & 0.8785       \\
\hline
BERT-base+ResNet18        & 0.8262\textcolor{blue}{$\uparrow$}       & 0.8263\textcolor{blue}{$\uparrow$}        & 0.8140       & 0.8739       \\
BERT-base+ResNet50        & \textbf{0.8266}\textcolor{blue}{$\uparrow$}        & \textbf{0.8297}\textcolor{blue}{$\uparrow$}        & \textbf{0.8251}\textcolor{blue}{$\uparrow$}        & 0.8821       \\
BERT-base+InceptionV4     & 0.8000       & 0.8058       & 0.8189\textcolor{blue}{$\uparrow$}        & 0.8873\textcolor{blue}{$\uparrow$}        \\
\hline
BERT-large+ResNet18       & 0.8153       & 0.8181       & 0.8128       & 0.8798\textcolor{blue}{$\uparrow$}        \\
BERT-large+ResNet50       & 0.8245\textcolor{blue}{$\uparrow$}        & 0.8267\textcolor{blue}{$\uparrow$}        & 0.8201       & 0.8817       \\
BERT-large+InceptionV4    & 0.8226\textcolor{blue}{$\uparrow$}       & 0.8289\textcolor{blue}{$\uparrow$}        & 0.8235\textcolor{blue}{$\uparrow$}        & \textbf{0.8884}\textcolor{blue}{$\uparrow$}        \\
\bottomrule
\end{tabular}
\label{tab:sub_table1}
\end{subtable}

\begin{subtable}{1.\linewidth}
\centering
\caption{Comparisons of performance with baseline on the AIGCIQA2023 database.}
\begin{tabular}{@{}l|cc|cc|cc@{}}
\toprule
 \multirow{2}{*}{Method} & \multicolumn{2}{c|}{Quality} & \multicolumn{2}{c|}{Authenticity} & \multicolumn{2}{c}{Correspondence} \\ \cline{2-7}  
 & SRCC & PLCC  & SRCC & PLCC & SRCC & PLCC  \\ 
\hline
ResNet18(Baseline)               & 0.8005 & 0.8220 & 0.7259 & 0.7197 & 0.6997 & 0.6953 \\
ResNet50(Baseline)                & 0.8179 & 0.8351 & 0.7307 & 0.7269 & 0.7172 & 0.7050 \\
InceptionV4(Baseline)              & 0.8265 & 0.8422 & 0.7430 & 0.7245 & 0.7109 & 0.7030 \\
\hline
BERT-base+ResNet18        & 0.8097\textcolor{blue}{$\uparrow$} & 0.8316\textcolor{blue}{$\uparrow$} & 0.7197 & 0.7131 & 0.7115\textcolor{blue}{$\uparrow$} & 0.6964\textcolor{blue}{$\uparrow$} \\
BERT-base+ResNet50        & 0.8194\textcolor{blue}{$\uparrow$} & 0.8359\textcolor{blue}{$\uparrow$} & 0.7323\textcolor{blue}{$\uparrow$} & 0.7226 & 0.7033 & 0.6966 \\
BERT-base+InceptionV4     & 0.8278\textcolor{blue}{$\uparrow$} & 0.8472\textcolor{blue}{$\uparrow$} & \textbf{0.7498}\textcolor{blue}{$\uparrow$} & \textbf{0.7420}\textcolor{blue}{$\uparrow$} & \textbf{0.7218}\textcolor{blue}{$\uparrow$} & \textbf{0.7173}\textcolor{blue}{$\uparrow$} \\
\hline
BERT-large+ResNet18       & 0.8131\textcolor{blue}{$\uparrow$} & 0.8318\textcolor{blue}{$\uparrow$} & 0.7234 & 0.7115 & 0.7000\textcolor{blue}{$\uparrow$} & 0.6921 \\
BERT-large+ResNet50       & 0.8172 & 0.8305 & 0.7354\textcolor{blue}{$\uparrow$} & 0.7337\textcolor{blue}{$\uparrow$} & 0.7047 & 0.6964 \\
BERT-large+InceptionV4    & \textbf{0.8326}\textcolor{blue}{$\uparrow$} & \textbf{0.8526}\textcolor{blue}{$\uparrow$} & 0.7485\textcolor{blue}{$\uparrow$} & 0.7403\textcolor{blue}{$\uparrow$} & 0.7210\textcolor{blue}{$\uparrow$} & 0.7148\textcolor{blue}{$\uparrow$} \\
\bottomrule
\end{tabular}

\label{tab:sub_table3}
\end{subtable}%
\end{table}

\subsection{Implementation Details}
Our experiments are conducted on the NVIDIA A40, using PyTorch 1.11.0 and CUDA 11.3 for both training and testing.

During training, the batch size $B$ is set to $8$. We utilize the Adam optimizer\cite{kingma2014adam} with a learning rate of $1 \times 10^{-4}$ and weight decay of $1 \times 10^{-5}$. The training loss employed is mean squared error (MSE) loss. In the testing phase, the batch size $B$ is set to $20$.

We choose the NR-AIGCIQA method\cite{yuan2023pkui2iqa} as the Baseline and report the performance of our proposed TIER method utilizing different Text encoders and Image encoders. They are as follows:

$\bullet$  \bm{$F^I+R$} (Baseline) : Corresponds to the NR-AIGCIQA method\cite{yuan2023pkui2iqa}. Here, $F^I$ represents the Image encoder and $R$ represents the Regression network.

$\bullet$  \bm{$F^I+F^T+R$} : Corresponds to the TIER method proposed in this paper. This method adds a Text encoder to the Baseline to extract text features from text prompts, which are then fused with the image features extracted by the Image encoder. Here, $F^I$ represents the Image encoder, $F^T$ represents the Text encoder, and $R$ represents the Regression network.

\subsection{Results and Analysis}
The performance results of the proposed methods utilizing different Text encoders and Image encoders on three mainstream AIGCIQA databases are exhibited in Table 1. We further provide a comparison as shown in Fig.3. Based on the results reported in Table 1, we can draw several conclusions:

$\bullet$  The experimental results on three mainstream AIGCIQA databases indicate that our proposed TIER method generally demonstrates superior performance compared to baseline in most cases.

$\bullet$  The experimental results of correspondence score on AIGCIQA2023 database indicate that despite our method considering both generated images and their corresponding text prompts in predicting correspondence scores, this does not necessarily lead to performance improvement. This suggests that our method may not fully comprehend the relationship between generated images and their corresponding text prompts in certain cases, thereby limiting its performance in predicting correspondence scores.

\section{Conclusion}
In this paper, we propose a text-image encoder-based regression (TIER) framework to address the issue that previous AIGCIQA methods often overlook the information contained in the text prompts of those generated images. To demonstrate the effectiveness of our proposed method, we conduct extensive experiments on three mainstream AIGCIQA databases. The experimental results indicate that our proposed TIER method generally demonstrates superior performance compared to baseline in most cases. We hope our methodology offers insights for future AIGCIQA research.

\clearpage
\bibliographystyle{plain}   
\bibliography{TIER} 
\end{document}